\documentclass[10pt,twocolumn,letterpaper]{article}

\usepackage{cvpr}              
\definecolor{cvprblue}{rgb}{0.21,0.49,0.74}
\usepackage[pagebackref,breaklinks,colorlinks,allcolors=cvprblue]{hyperref}
\usepackage{multicol}
\usepackage{multirow}
\usepackage{float}
\usepackage{gensymb}
\usepackage{pifont}
\usepackage{algorithm}
\usepackage{algorithmic}

\usepackage{listings}
\usepackage[utf8]{inputenc}
\usepackage[T1]{fontenc}
\usepackage[accsupp]{axessibility}  

\def\paperID{4041} 
\def\confName{CVPR}
\def\confYear{2026}

\title{Revisiting Articulated Parts Perception in Robot Manipulation}

\author{
Xiaoqian Wu,~
Yejie Guo,~
Xiaoyang Chen,~
Lixin Yang,~
Cewu Lu$^*$,~
Yong-Lu Li$^*$\\
{\ttfamily\small Shanghai Jiao Tong University}\\ 
{\ttfamily\small \{enlighten, gyj123, cxy\_computer, siriusyang, lucewu,  yonglu\_li\}@sjtu.edu.cn}
}

\begin{document}

\begin{figure}
\twocolumn[{%
\renewcommand\twocolumn[1][]{#1}
\maketitle
\centering
\includegraphics[width=1\textwidth]{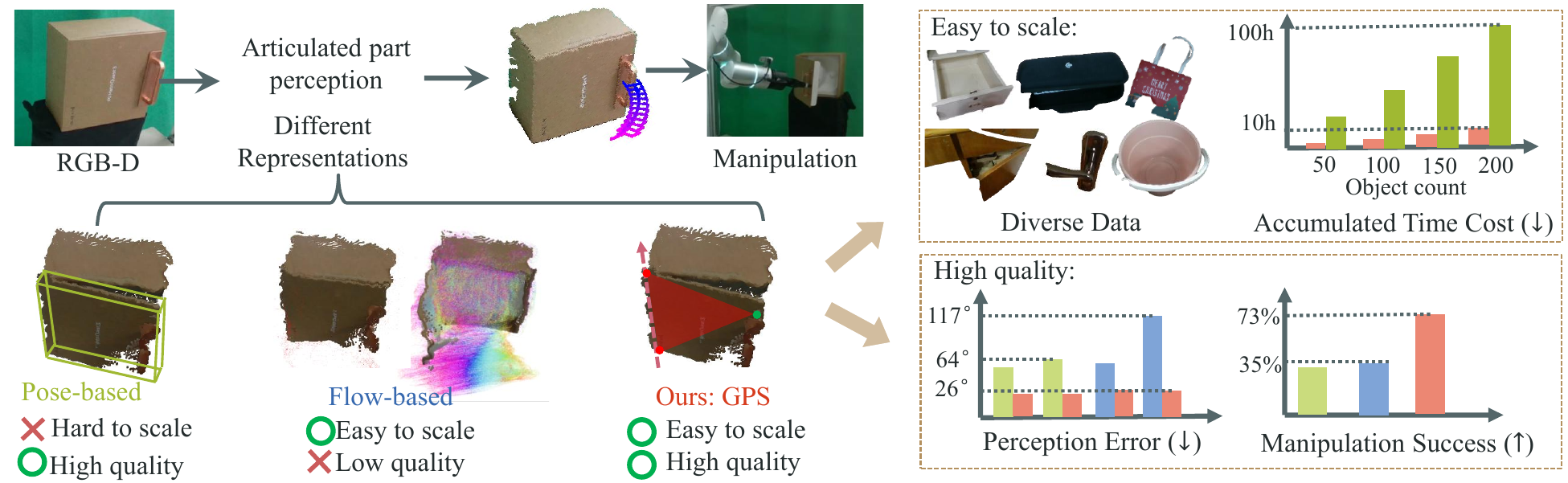}
\captionof{figure}{\textbf{Overview.} We aim to enhance robotic manipulation by improving articulated part perception from a single RGB-D image. 
The core of our approach is a novel affordance representation, GPS, which is easy to scale with high-quality data.
Our model outperforms existing pose-based and flow-based methods in part perception accuracy and manipulation success rate. 
}
\vspace{1.0cm}
\label{fig:teaser} 
}]
\end{figure}

\let\thefootnote\relax\footnotetext{$^*$Corresponding author.}

\begin{abstract}

We are surrounded by various objects with movable, articulated parts, \eg, box, handle, door.  
An accurate and generalizable perception of articulated parts is essential to enhance robotic manipulation capabilities.
Building on this need, recent efforts in articulated parts perception have followed two main directions: One line of work uses pose-based representation, which requires high manual cost; in parallel, affordance-based methods extract future object motion from point tracking without additional manual efforts, but suffer from low-quality data.
In this paper, we propose a new representation of articulated parts, \textbf{G}eometric \textbf{P}rimary  \textbf{S}tructure \textbf{(GPS)}, an abstraction of the part geometry structure to balance scalability and quality.
For efficient and scalable data collection, GPS is integrated with a portable Virtual Reality (VR) device and requires only one minute to annotate one object sequence.
This direct human annotation provides higher quality than the estimated affordance.
With this efficient VR-GPS system, we collect \textbf{41K} frames for \textbf{234} objects across six part classes, and train a generalizable GPS model with a single RGB-D object image as input. 
For object manipulation, we deploy a heuristic policy based on GPS prediction. Without any in-domain fine-tuning, our method achieves an \textbf{73\%} success rate, covering 270 initial states for 9 objects.
Our code, data and reusable tool are available at \href{https://enlighten0707.github.io/gps}{https://enlighten0707.github.io/gps}.
\end{abstract}

\section{Introduction}
\label{sec:intro}


Accurately estimating object state is crucial for robots to perform diverse manipulation tasks.
While recent advances have improved state estimation for rigid objects~\cite{onepose++,foundationpose}, non-rigid objects such as articulated objects remain challenging due to their complex kinematic structures.
In this paper, we focus on enhancing the perception and estimation of articulated parts from a single RGB-D image to improve  robotic manipulation capabilities, as illustrated in Fig.~\ref{fig:teaser}.

Recent efforts in articulated parts perception have followed two main directions: pose-based and affordance-based representations.
\textit{Pose-based} methods represent parts as segmentation and pose estimation, defining canonical positions and orientations for each part class~\cite{geng2023gapartnet, npcs}.
However, obtaining such data requires significant manual effort: synthetic CAD models are created by professional artists~\cite{xiang2020sapien,mo2019partnet,akb48}, and real-world objects require elaborate scanning, modeling, and frame-wise pose annotation~\cite{liu2022hoi4d,zhan2024oakink2,liu2024taco}.
Although post-processing methods have emerged to reconstruct articulated objects from visual inputs~\cite{ditto,rsrd,artgs,paris}, they still face limitations including category restriction~\cite{ditto}, long processing time~\cite{rsrd}, and sensitivity to errors in real-world scenarios~\cite{artgs,paris}.

In parallel, \textit{affordance-based} methods~\cite{wen2023any,yuan2024general,bahl2023affordances} model object motion by predicting future point trajectories, often referred to as object flow. 
The ground truth flow is typically extracted automatically from human-object interaction videos via point tracking followed by temporal down-sampling. 
While this minimizes manual annotation, it suffers from two key issues:
\textbf{1) Tracking Error.} Point tracking~\cite{xiao2025spatialtrackerv2,trace-anything} is prone to inaccuracies caused by self-occlusion or camera movement.  
In Fig.~\ref{fig:teaser}, 
when opening the door, the trajectories near the edge deviate significantly from the true motion, and those near the axis should not have been truncated so prematurely;
\textbf{2) Flow Ambiguity.} The mapping from point trajectories to articulated properties is inherently ambiguous. 
The scale of the extracted flow varies for objects of similar type but different sizes, and sub-flows are temporally inconsistent for manipulations with the same range of motion but non-uniform execution speed. This makes the predicted flow highly sensitive to both object scale and temporal variations.

\begin{figure}[t]
\begin{center}
\includegraphics[width=0.8\linewidth]{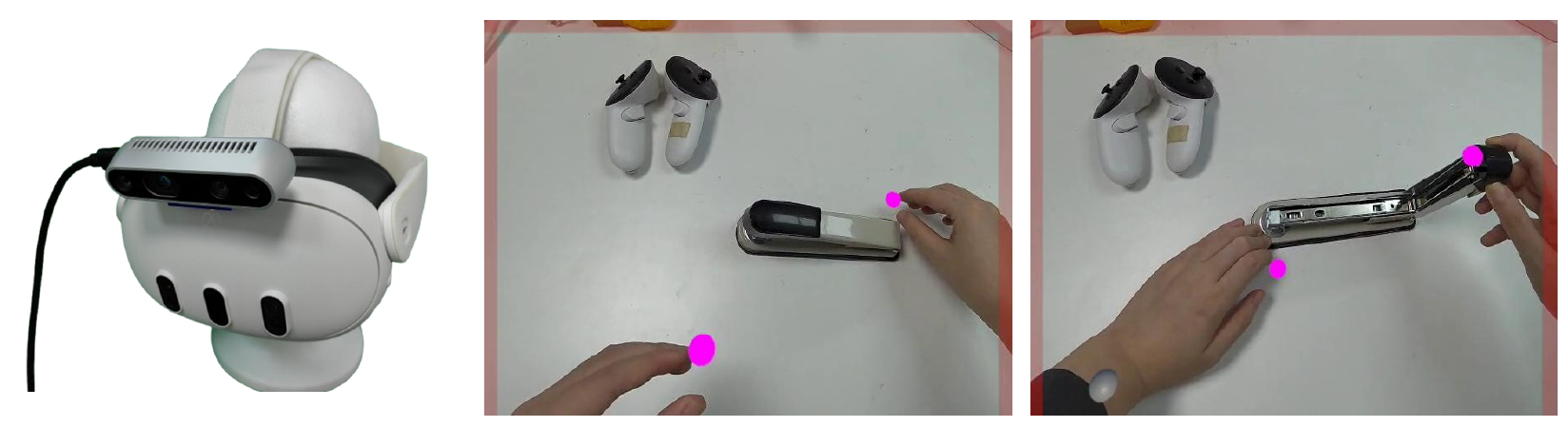}
\caption{\textbf{VR hardware and interfaces.} The headset tracks the fingers and renders the tracking points as red points. 
}
\label{fig:vr_interface}
\end{center}
\vspace{-15pt}
\end{figure}

To address these limitations, we need an abstraction of the part geometry that is both informative and scalable. Our solution comprises two key components:
\textbf{1) Explicit Axis Annotation.} 
We propose to explicitly annotate and predict the motion axis, \eg,  a revolute axis for a laptop lid, a prismatic axis for a drawer. 
This axis captures  an inherent invariance during part motion, which helps to mitigate noise from point tracking and the ambiguity in flow.
To enable accurate and efficient annotation,  we employ a Virtual Reality (VR) device, Meta Quest 3, with SLAM capabilities. 
Before interacting  with the object, virtual points are placed on the axis position.
These points remain stationary in 3D space and are unaffected by headset movement, and are rendered in the headset for annotators to verify and adjust.
\textbf{2) Hand as a Motion Proxy.}
Instead of tracking the object part directly, which can be unreliable, we use the human hand firmly grasping the part as a stable motion proxy. 
The hand has a consistent structure and is typically closer to the camera, making it more suitable for robust tracking than the object part.  
We track the midpoint between the thumb and index finger to represent the part's motion, and this hand point is also visually rendered in the headset. Example interfaces are shown in Fig.~\ref{fig:vr_interface}.
Leveraging this efficient data collection method, we introduce the \textbf{G}eometric \textbf{P}rimary  \textbf{S}tructure \textbf{(GPS)} as a new affordance representation.
As is shown in Fig.~\ref{fig:teaser}, GPS is defined by three keypoints: two (red) determine the axis, and a third (green) is attached to the hand. The three keypoints form a plane (red), constraining the part rotation. 
Part translation can be formulated similarly, constrained by a prismatic axis and a plane perpendicular to it.

Our VR-GPS system requires only one minute to annotate a hand-object interaction video, without any manual post-processing, and yields higher-quality data than estimated flow.
Using this system, we have collected a dataset including \textbf{41K} RGB-D frames for \textbf{234} objects across six part classes, providing rich object knowledge.
Based on this data, we propose a generalized GPS prediction model that transfers well to other datasets and outperforms both pose-based and flow-based methods in articulated parts understanding.
As the first step, we verify the effectiveness of our data and hope the community will join us to enlarge our dataset continuously with the reusable VR-GPS system.

For robot manipulation, we develop a heuristic policy, using the predicted GPS to select initial grasp proposals from AnyGrasp~\cite{fang2023anygrasp} and generate subsequent  waypoints. 
Real-robot experiments cover 270 initial states (\ie, different part poses and camera views) for 9 objects with diverse appearances. 
Without any in-domain fine-tuning, our method achieves an impressive 73\% success rate.

Our main contributions are:
1) The novel GPS representation for articulated parts, which balances between scalability and quality in data collection.  
2) A VR-collected GPS dataset rich in object geometry knowledge. 
3) A generalizable GPS prediction model is proposed that demonstrates superior performance in facilitating real-world robotic manipulation of daily objects.


\section{Related Work}
To enhance robot manipulation, a system should understand both interaction semantics~\cite{shan2020understanding, grauman2022ego4d,wu2023symbol} and geometry of manipulated objects. In this paper, we focus on understanding articulated object geometry. Prior work on articulated object representation can be broadly grouped into two main categories: pose-based and affordance-based methods.

\textbf{Pose-based methods} 
aim to estimate part segmentation and 6-DoF pose, often defined in a Normalized Part Coordinate Space (NPCS) for each object category~\cite{nocs,npcs}. 
GAPartNet~\cite{geng2023gapartnet} extends this idea by introducing cross-category part classes based on functional similarity.
Data for pose-based methods primarily come from three sources:
1) Synthetic datasets, which provide high-fidelity 3D assets created by artists in simulation platforms~\cite{makoviychuk2021isaac,mo2019partnet,xiang2020sapien,shen2021igibson,geng2023gapartnet}; 
2) Real-world scans, involving professionally captured object models~\cite{calli2015ycb,hinterstoisser2011multimodal,martin2019rbo,akb48} with frame-wise pose and camera annotations from hand-object interactions~\cite{liu2022hoi4d,zhan2024oakink2,liu2024taco};
3) Post-processing methods that recover articulation from visual inputs.
For example, 
PARIS~\cite{paris} and ArtGS~\cite{artgs} leverages  neural radiance fields and 3D Gaussians to reconstruct objects and estimate joints, and RSRD~\cite{rsrd} uses a 4D differentiable part model to recover object motions from an object scan and single monocular video.

\textbf{Affordance-based methods}
focus on identifying \textit{where} to manipulate (\ie, contact point) an object and \textit{how} to manipulate (\ie, future trajectory).
Contact point prediction are often predicted from annotated point clouds~\cite{affordancenet} or semantic features from diffusion models~\cite{ju2024robo}.
Beyond contact points, predicting the future trajectory of an object is also crucial for robot manipulation.
VRB~\cite{bahl2023affordances} derive affordance cues from hand-object contact and hand motion, employing 2D representations as guidance for robot learning. However, 2D affordance offers only coarse supervision.
GFlow~\cite{yuan2024general} extracts future object motion from point tracking on the HOI4D dataset~\cite{liu2022hoi4d}, assuming known camera parameters for each frame. Yet, in practice, most real-world data lacks precisely annotated camera parameters and consists only of visual inputs.
While recent methods improve dynamic scene reconstruction from monocular views~\cite{wang2024shape,zhang2024monst3r,trace-anything}, they still struggle with articulated objects and remain error-prone.
\section{Definition}

\begin{figure}[t]
\begin{center}
\includegraphics[width=1\linewidth]{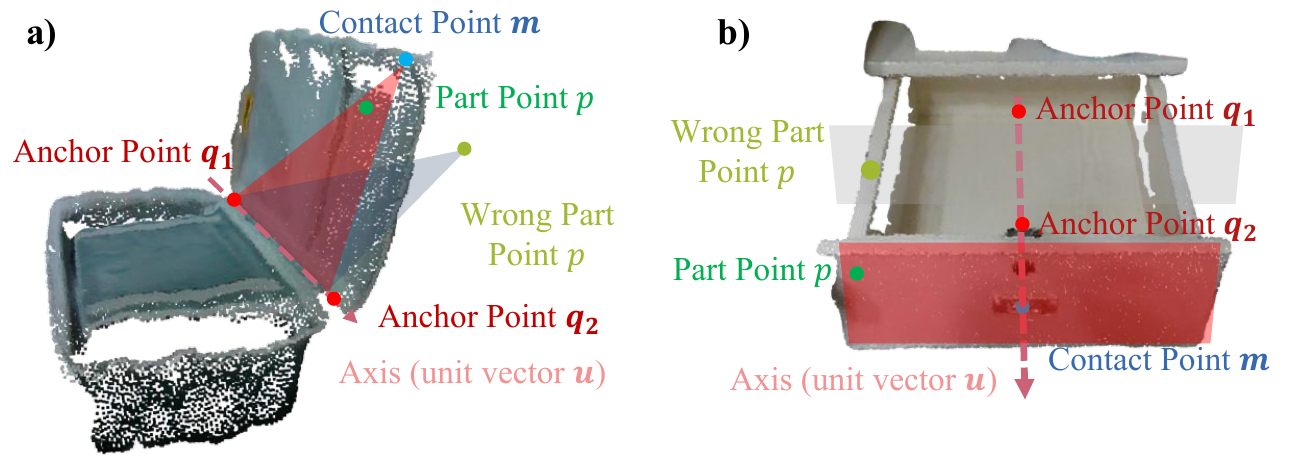}
\caption{\textbf{Geometric structure formulation.} (a) Part rotation along revolute axis; (b) part translation along prismatic axis.}
\label{fig:kp_def}
\end{center}
\vspace{-20pt}
\end{figure}

\subsection{Part Rotation}
\label{sec:revolute}
Many real-world objects have rotatable parts, \eg, box lid, kettle, book, lamp. 
Given an input point cloud ${\mathcal{P} \in \mathbb{R} ^{N\times 3}}$ of an object, its structure is constrained as $\{\mathbf{u}, \mathbf{q}, \mathbf{m}\}$. Here $\mathbf{u}\in \mathbb{R}^3$ is a unit vector of the revolute axis direction, $\mathbf{q}$ is an anchor point to determine the position of the evolute axis, and $\mathbf{m}$ is a contact point where a movable part contacts with a human hand or a robot end-effector.
The trajectory of $\mathbf{m}$ concerning a rotation angle 
$\theta$ is
\begin{equation}
    \mathbf{m}(\theta) = \cos(\theta) I \cdot \mathbf{m} + (1 - \cos(\theta)) \mathbf{u} \mathbf{u}^T \cdot \mathbf{m} + \sin(\theta) \mathbf{R} \cdot \mathbf{m} + \mathbf{q}
    \label{eq:revolute_motion},
\end{equation}
where $I$ denotes an identity matrix and $R$ denotes the skew symmetric matrix of $\mathbf{u}$.

Our GPS is defined as $\{\mathbf{q}_1, \mathbf{q}_2, \mathbf{p}\}$ (Fig.~\ref{fig:kp_def}(a)).
The axis $\{\mathbf{u}, \mathbf{q}\}$ is determined by two anchor points $\{\mathbf{q}_1, \mathbf{q}_2\}$, where $\mathbf{q}_1=\mathbf{q}$, $\mathbf{q}_2=\mathbf{q}+c \mathbf{u}$, $c$ is an arbitrary constant. 
The contact point $\mathbf{m}$ is not unique. For example, when opening the bag in Fig.~\ref{fig:kp_def}(a), we can touch different positions on three edges. Thus, learning the exact position of a contact point will increase the difficulty of GPS learning. Also, the predicted $\mathbf{m}$ cannot be benchmarked accurately. 
Therefore, we define a part point $\mathbf{p}$ as:
\begin{equation}
    \mathbf{p} \cdot ((\mathbf{q}_1 - \mathbf{m}) \times (\mathbf{q}_2 - \mathbf{m})) = 0,
\end{equation}
$\mathbf{p}$ should be on the plane defined by $\mathbf{q}_1, \mathbf{q}_2, \mathbf{m}$, constraining the principal structure. 
Additionally, in robot manipulation experiments (Sec.~\ref{sec:robot}), we conduct an ablation study and find that using loose constraints $\mathbf{p}$ is more generalized than $\mathbf{m}$ to select better grasp proposals.

\subsection{Part Translation}
\label{sec:prism}
Other objects have parts that translate along a prismatic axis.
Its structure is constrained by $\{\mathbf{u}, \mathbf{m}\}$, where $\mathbf{u}$ is a unit vector of the axis direction,  $\mathbf{m}$ is the contact point (Fig.~\ref{fig:kp_def}(b)). The trajectory of $\mathbf{m}$ with respect to an offset $\delta$ is:
\begin{equation}
    \mathbf{m}(\theta) = \mathbf{m} + \delta \mathbf{u},
    \label{eq:prism_motion}
\end{equation}
Its GPS is $\{\mathbf{q}_1, \mathbf{q}_2, \mathbf{p}\}$, where $\{\mathbf{q}_1, \mathbf{q}_2\}$ defines axis direction: $\mathbf{u} \cdot (\mathbf{q}_1 - \mathbf{q}_2)=0$. 
For translational parts, the axis defines the direction of motion but not its absolute position. To simplify model learning, we fix this axis to pass through the part's geometric center.
As is demonstrated in Sec.~\ref{sec:revolute}, the contact point $\mathbf{m}$ is loosed to part point $\mathbf{p}$:
\begin{equation}
    (\mathbf{p} - \mathbf{m}) \cdot (\mathbf{q}_1 - \mathbf{q}_2) = 0,
\end{equation}
$\mathbf{p}$ should be on the plane defined by normal vector $\mathbf{q}_1 - \mathbf{q}_2$ and contact point $\mathbf{m}$.

\section{Data Collection}

\begin{figure*}[t]
\begin{center}
\includegraphics[width=0.85\linewidth]{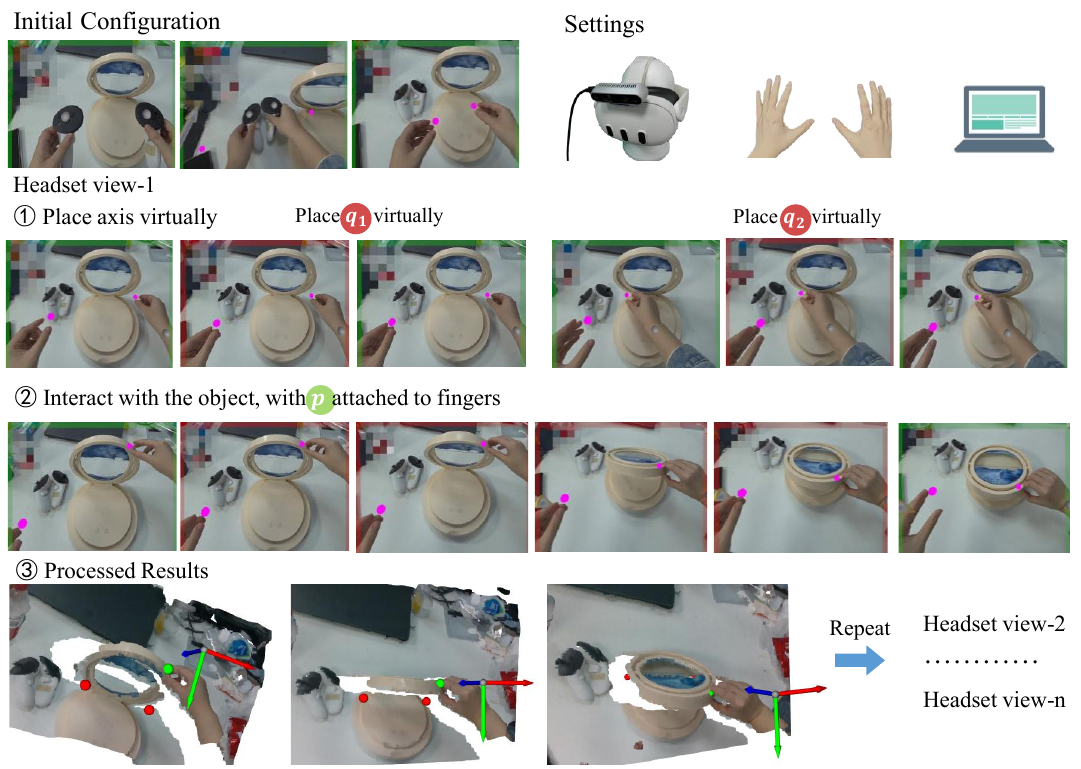}
\caption{\textbf{Hardware settings and annotation pipeline.}
Before interacting with an object, the annotator places axis $\{\mathbf{q_1}, \mathbf{q_2}\}$ virtually. During interaction, the part point $\mathbf{p}$ is attached with fingers. For each object, multiple RGB-D videos with different headset views are recorded. 
The annotator begins and ends the recording by performing a pinch gesture with their non-interacting hand.
}
\label{fig:ar_pipeline}
\end{center}
\vspace{-10pt}
\end{figure*}

\subsection{System Design}
Our system is built around the Meta Quest 3 VR device.
The system consists of the following components:
\textbf{1) VR headset} serving as a display and providing spatial computation. Based on its Augmented Reality (AR) functionality, we can virtually place a point in the real 3D world, track its movement, and record its coordinates in the scene point cloud. 
Hand tracking is performed using the built-in VR function, with the midpoint between the thumb and index finger being tracked and rendered in real time within the headset interface.
\textbf{2) Intel RealSense D435 camera} is mounted on the headset via a 3D-printed bracket to capture RGB-D data and reconstruct the scene point cloud~\cite{chen2024arcap}. 
The virtual point is defined in the world frame of the VR device, while the scene point cloud is defined in the RealSense camera frame. Therefore, after recording, the coordinate of the virtual point is transformed into the camera frame, using calibration parameters between RealSense and the headset. 
\textbf{3) Laptop} receiving and storing data streams from both the VR device and the RealSense camera.

We design the annotation process to record GPS in real time without requiring complex post-processing.
To annotate the axis points $\{\mathbf{q_1}, \mathbf{q_2}\}$,  the annotator places virtual points along  the axis \textit{before} interaction. 
Then, the points stay \textit{fixed} in the current 3D space despite headset move and camera view move, with the help of spatial memory in VR computing. 
To annotate the part point $\mathbf{p}$, the annotator interacts with the object while recording RGB-D video, with the virtual points \textit{moving} with the annotator's fingers.

The overall annotation pipeline is shown in Fig~\ref{fig:ar_pipeline}.
After initial configuration, the annotator annotates each object sequentially. 
For each object, the annotator places axis points $\{\mathbf{q_1}, \mathbf{q_2}\}$ virtually records RGB-D videos, then changes to another headset view and repeats the process.
After recording, we apply coordinate transformation and object segmentation~\cite{ravi2024sam2}. 
Finally, we obtain the object RGB-D data with GPS annotation across different object part poses and camera views.

One unique advantage of AR-based annotation is that one point can be placed \textit{anywhere} in the 3d space. 
To annotate points, a direct way is to click on pixels after recording RGB-D videos and map pixel coordinates to 3D points based on the camera intrinsic. 
However, when the correct points are not on the surface facing the camera, \eg, the revolute axis of a thin lamp, it is difficult to accurately annotate them.
Moreover, the real-time visibility of virtual points allows annotators to adjust and correct placements during interaction, which is an advantage over offline methods such as hand reconstruction or 3D GUI-based annotation.
Furthermore, built on a widely available commercial VR device, our VR-GPS is portable, not limited to lab settings.


\subsection{Data Analysis}
{\bf Low Cost.}
The device cost of our VR-GPS is relatively low (\textit{800 dollars}), 
without an expensive MoCap system or 3D scanning devices.
For each object, three videos with different camera views are recorded.
The average time to annotate one video is one minute, which is efficient.

\begin{figure}[t]
\begin{center}
\includegraphics[width=1.0\columnwidth]{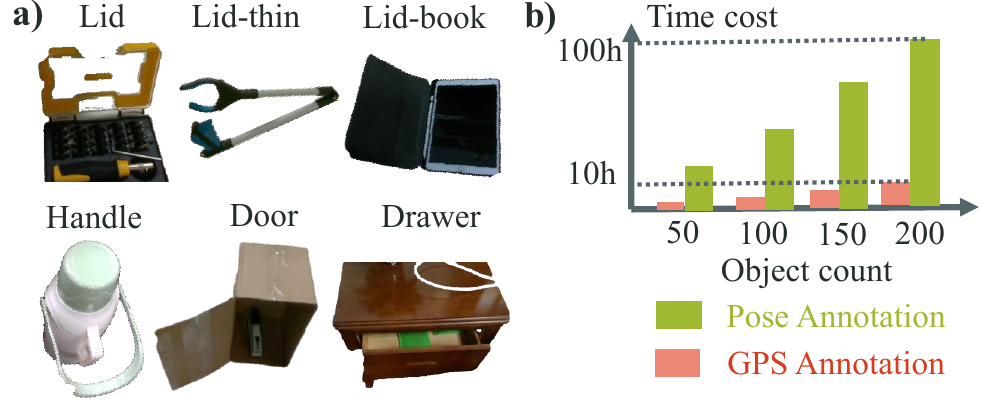}
\caption{\textbf{Dataset overview.} Our VR-GPS is diverse and efficient.}
\label{fig:dataset}
\end{center}
\vspace{-5pt}
\end{figure}

\noindent{\bf Dataset Statistics.}
Using our portable and efficient VR-GPS, we collect 41K frames for 234 objects. As shown in Fig.~\ref{fig:dataset}(a), the objects belong to six part classes: Lid (\eg, Box, Laptop), Lid-thin (\eg, Pole, Stapler), Lid-book (\eg, Ipad, Booklet), Handle (\eg, Kettle, Bucket), Door (\eg, Safe, Cabinet), Drawer. The drawer has a  prismatic axis, while the others have a  revolute axis.
Fig.~\ref{fig:dataset}(b) compares the time cost of GPS annotation against pose-based annotation, highlighting the efficiency of our approach.

\noindent{\bf Data Quality.}
The error of the virtual point coordinate mainly comes from poor lighting conditions, or
too small a distance between the annotator and the VR-defined boundary.
We double-check the collected data to ensure quality. After checking, 3\% of the data is filtered.


\section{Geometric Structure Learning}



\subsection{Model Design}

\textbf{Feature Extraction.}
Given RGB image $\mathcal{I} \in \mathbb{R} ^{H\times W \times 3}$, the depth map $\mathcal{D} \in \mathbb{R} ^{H\times W}$,
our goal is to predict GPS parameters $\{\mathbf{q}_1, \mathbf{q}_2, \mathbf{p}\}$.
Following CAPNet~\cite{capnet}, we use SAM2~\cite{ravi2024sam2} and FeatUp~\cite{featup} to extract RGB feature maps, where each pixel corresponds to a vector with dimension $d_s=480$ representing the semantic information of the RGB image at the corresponding location.
Subsequently, we concatenate each RGB feature vector
with its corresponding 3D point cloud $\mathcal{P} \in \mathbb{R}^3$ in a point-wise manner.
To aggregate part category semantics, we use CLIP~\cite{clip} text encoder to convert category descriptions into semantic features. Then their dimensions are reduced via MLPs to $d_t=6$, and concatenated with $\mathcal{P}$.
Finally, the merged features ${\mathcal{P}_t \in \mathbb{R} ^{N\times (6+d_s+d_t)}}$ are processed via PointNet++~\cite{qi2017pointnet++} to extract the densely fused RGBD features $f$.

\noindent{\bf{Part Rotation Loss Design.}}
For part rotation along revolute axis, the geometric features $f$ is processed via three separate MLPs to predict axis direction $\hat{\mathbf{u}}$, the axis anchor point $\hat{\mathbf{q}_1}$, and the part point $\hat{\mathbf{p}}$. 
The training loss is
$\mathcal{L} = \mathcal{L}_{ad} + \mathcal{L}_{ao} + \mathcal{L}_{pd}$, referring to the axis direction loss, axis offset loss, and part direction loss:
\begin{align}
     \mathcal{L}_{ad} & = 1 - \left\vert \frac{\hat{\mathbf{u}}}{\Vert\hat{\mathbf{u}}\Vert} \cdot \frac{\mathbf{q}_1-\mathbf{q}_2}{\Vert\mathbf{q}_1-\mathbf{q}_2\Vert} \right\vert \label{eq:L_ad},\\
     \mathcal{L}_{ao} &= \left\vert \frac{\hat{\mathbf{q}_1} - \mathbf{q}_1}{\Vert\hat{\mathbf{q}_1} - \mathbf{q}_1\Vert} \cdot \frac{\mathbf{q}_2-\mathbf{q}_1}{\Vert\mathbf{q}_2-\mathbf{q}_1\Vert} \right\vert,
     \label{eq:L_ao},\\
     \mathcal{L}_{pd} & = 1 - \frac{(\mathbf{q}_2-\mathbf{q}_1)\times \mathbf{p}}{\Vert(\mathbf{q}_2-\mathbf{q}_1)\times \mathbf{p}\Vert} \cdot \frac{(\mathbf{q}_2-\mathbf{q}_1)\times \hat{\mathbf{p}}}{\Vert(\mathbf{q}_2-\mathbf{q}_1)\times \hat{\mathbf{p}}\Vert},
\end{align}

\noindent{\bf{Part Translation Loss Design.}}
For part translation along the prismatic axis, 
the training loss is
$\mathcal{L} = \mathcal{L}_{ad} + \mathcal{L}_{ao} + \mathcal{L}_{po}$, 
referring to the axis direction loss, the axis offset loss and
the part offset loss $\mathcal{L}_{po}$:
\begin{align} 
    \mathcal{L}_{po} = \frac{\vert((\mathbf{q}_2-\mathbf{q}_1) \times (\mathbf{p}-\mathbf{q}_1))\cdot(\hat{\mathbf{p}}-\mathbf{q}_1)\vert}{\Vert(\mathbf{q}_2-\mathbf{q}_1) \times (\mathbf{p}-\mathbf{q}_1)\Vert}. 
\end{align}

\begin{table}[t]
        \centering
        \resizebox{0.9\linewidth}{!}{
        \begin{tabular}{l |l | ccc}
            \toprule
            \textbf{Category} & \textbf{Method} & \textbf{AADE} ($\downarrow$) & \textbf{AAOE}($\downarrow$) & \textbf{APDE}($\downarrow$)\\
            \midrule
            \multirow{3}{*}{Laptop} & CAPNet~\cite{capnet} &37.33\degree & 0.33 & 62.78\degree \\
            & Ours-Flow & 23.98\degree & 0.13 & 16.78\degree\\
             & \textbf{Ours} & \textbf{17.17}\degree & \textbf{0.13} & \textbf{9.26}\degree \\
             \midrule
            \multirow{3}{*}{Trashcan} & CAPNet~\cite{capnet} &34.54\degree & 0.28 & 46.88\degree \\
            & Ours-Flow & 29.57\degree & 0.25 & 28.93\degree\\
             & \textbf{Ours} & \textbf{24.46}\degree & \textbf{0.21} & \textbf{18.27}\degree \\
             \midrule
            \multirow{3}{*}{Safe} & CAPNet~\cite{capnet} & 56.49\degree & 0.48 & 85.10\degree\\
            & Ours-Flow & 27.51\degree & 0.38 & 51.36\degree\\
             & \textbf{Ours} & \textbf{15.52}\degree & \textbf{0.26} & \textbf{33.75}\degree\\
             \midrule
            \multirow{3}{*}{Bucket} & CAPNet~\cite{capnet} &48.21\degree & 0.41& 61.70\degree \\
            & Ours-Flow & 39.05\degree & 0.31 & 28.28\degree\\
             & \textbf{Ours} & \textbf{32.05}\degree & \textbf{0.27}& \textbf{19.60}\degree \\
             \midrule
             \midrule
             \textbf{Category} & \textbf{Method} & \textbf{AADE} ($\downarrow$) & \textbf{AAOE}($\downarrow$) & \textbf{APOE}($\downarrow$)\\
             \midrule
             \multirow{3}{*}{Drawer} & CAPNet~\cite{capnet} & 37.82\degree & 0.13 & 0.12\\
             & Ours-Flow & 17.30\degree & 0.26 & 0.16\\
             & \textbf{Ours-GPS} &  \textbf{14.32}\degree & \textbf{0.25} & \textbf{0.14}\\
            \bottomrule
        \end{tabular}}
        \caption{GPS learning performance on HOI4D~\cite{liu2022hoi4d} .}
        \label{tab:kp_hoi4d}
\end{table}

\begin{table}[t]
        \centering
        \resizebox{0.9\linewidth}{!}{
        \begin{tabular}{l |l | ccc}
            \toprule
            \textbf{Category} & \textbf{Method} & \textbf{AADE} ($\downarrow$) & \textbf{AAOE}($\downarrow$) & \textbf{APDE}($\downarrow$)\\
            \midrule
            \multirow{3}{*}{Laptop} & GFlow~\cite{yuan2024general} & 51.58\degree & 0.33 & 116.05\degree\\
            & Ours-Flow & 33.51\degree & 0.29 & 49.27\degree\\
             & \textbf{Ours} & \textbf{22.76}\degree & \textbf{0.19} & \textbf{22.81}\degree \\
             \midrule
            \multirow{3}{*}{Trashcan} & GFlow~\cite{yuan2024general}& 52.40\degree & 0.82 & 120.00\degree
             \\
             & Ours-Flow & 38.90\degree & 0.57 & 42.07\degree\\
             & \textbf{Ours} & \textbf{27.65}\degree & \textbf{0.27} & \textbf{20.79}\degree \\
             \midrule
            \multirow{3}{*}{Safe} & GFlow~\cite{yuan2024general} & 65.78\degree & 0.68 & 102.77\degree \\
            & Ours-Flow & 28.92\degree & 0.29 & 45.81\degree\\
             & \textbf{Ours} & \textbf{19.29}\degree & \textbf{0.23} & \textbf{22.96}\degree\\
             \midrule
            \multirow{3}{*}{Bucket} & GFlow~\cite{yuan2024general}  & 59.76\degree & 0.74 & 128.74\degree
            \\
            & Ours-Flow & 53.98\degree & 0.46 & 69.03\degree\\
             & \textbf{Ours} &  \textbf{36.25}\degree & \textbf{0.18} & \textbf{38.85}\degree\\
             \midrule
             \midrule
             \textbf{Category} & \textbf{Method} & \textbf{AADE} ($\downarrow$) & \textbf{AAOE}($\downarrow$) & \textbf{APOE}($\downarrow$)\\
             \midrule
             \multirow{3}{*}{Drawer} & GFlow~\cite{yuan2024general} & 61.83\degree & 0.81 & 0.98\\
             & Ours-Flow & 28.41\degree & 0.31 & 0.38\\
             & \textbf{Ours-GPS} &  \textbf{23.78}\degree & \textbf{0.20} & \textbf{0.21} \\
            \bottomrule
        \end{tabular}}
        \caption{GPS learning performance on RGBD-Art~\cite{capnet}.}
        \label{tab:kp_capnet}
\end{table}

\subsection{Evaluation}
\subsubsection{Benchmark}
{\textbf{Metrics.}}
To evaluate GPS prediction, we design metrics that quantify the direction and offset errors of both the axis and the part.
For the axis, we adopt metrics
Average Axis Direction Error (AADE) and Average Axis Offset Error (AAOE).
For the parts, we adopt Average Part Direction Error (APDE) for part rotation,
and Average Part Offset Error (APOE) for part translation. 
The maximum offset error is 2 with the point cloud normalized into a unit cube.

\noindent{\textbf{Test Datasets.}}
The GPS model is trained on our VR-GPS dataset. 
To assess its generalization capability, we evaluate the model on two external datasets: HOI4D~\cite{liu2022hoi4d} and RGBD-Art~\cite{capnet}.
\textbf{HOI4D} contains egocentric  RGB-D videos capturing human-object hand interactions.
\textbf{RGBD-Art} contains synthetic articulated objects annotations  built upon GAPartNet~\cite{geng2023gapartnet} dataset, featuring photorealistic RGB images and depth noise simulated like real sensors. 
They both provide ground-truth part segmentation and pose estimation, and we convert it into GPS by deriving $\mathbf{q}_1$, $\mathbf{q}_2$, and $\mathbf{p}$ from the bounding boxes.

\noindent{\bf{Test Object Categories.}}
We evaluate on five object categories: Laptop, Trashcan, Door, Bucket, and Drawer, the overlapping categories among VR-GPS, the test datasets, and the baselines. 
These categories correspond to the following parts: Laptop and Trashcan have a Lid, Safe has a Door, and Drawer has a Drawer.
The part classes Lid-thin and Lid-book are excluded due to a lack of suitable test data in HOI4D and RGBD-Art, where corresponding object categories are absent or interactions are largely pick-and-place with minimal articulation. 
These categories will instead be evaluated in the robot experiments in Sec.~\ref{sec:robot}.

\subsubsection{Performance Comparison}

We first evaluate our method against the pose-based method,
\textbf{CAPNet}~\cite{capnet}, predicting articulated part segmentation and pose, and trained on synthetic dataset RGBD-Art.
We convert its output into our GPS representation and evaluate on HOI4D, which serves as an out-of-domain benchmark for both methods.
We do not compare performance on RGBD-Art because it overlaps with CAPNet’s training domain.
As shown in Tab.~\ref{tab:kp_hoi4d}, GPS outperforms CAPNet across all five categories.
Although CAPNet employs depth noise augmentation, it still struggles with the sim-to-real gap, whereas our real-world data strategy alleviates this issue. 
We observe  a relatively higher GPS prediction error for the Bucket category, possibly because of the large depth sensor noise on its thin handle.

We next compare with the flow-based method \textbf{GFlow}~\cite{yuan2024general}, a state-of-the-art 3D scene flow predictor. 
We use its public checkpoint ScaleFlow-L trained on HOI4D and evaluate on the out-of-domain RGBD-Art benchmark.
The predicted flow is converted into GPS for direct comparison.
In Tab.~\ref{tab:kp_capnet}, GFlow exhibits large errors.
We attribute the results to the  inherent sensitivity of flow and the limited diversity of its training data, which hinders its ability to generalize to novel object instances.

We further fine-tune GFlow with our data (\textbf{Ours-Flow}). To adapt our data for flow-based training, we use TraceAnything~\cite{trace-anything} to extract globally aligned point trajectories under a moving camera. 
Following the setup in GFlow, each interaction sequence is divided into 4 timesteps. 
For a fair comparison with GPS, we modify the original GFlow by using object-level point clouds with annotated masks and performing per-point flow prediction.
Tab.~\ref{tab:kp_hoi4d} and \ref{tab:kp_capnet} show that Ours-GPS outperforms Ours-Flow because Ours-Flow suffers from inaccurate tracking results and inherent ambiguity of flow. 
Additionally, the performance of Ours-GPS degrades on RGBD-Art compared to HOI4D, primarily due to the simulated depth data in RGBD-Art lacking real-world fidelity. Nevertheless, we selected RGBD-Art for evaluation as it contains diverse articulated objects.

\section{Real Robot Experiments}
\label{sec:robot}

\begin{figure}[t]
\begin{center}
\includegraphics[width=0.99\columnwidth]{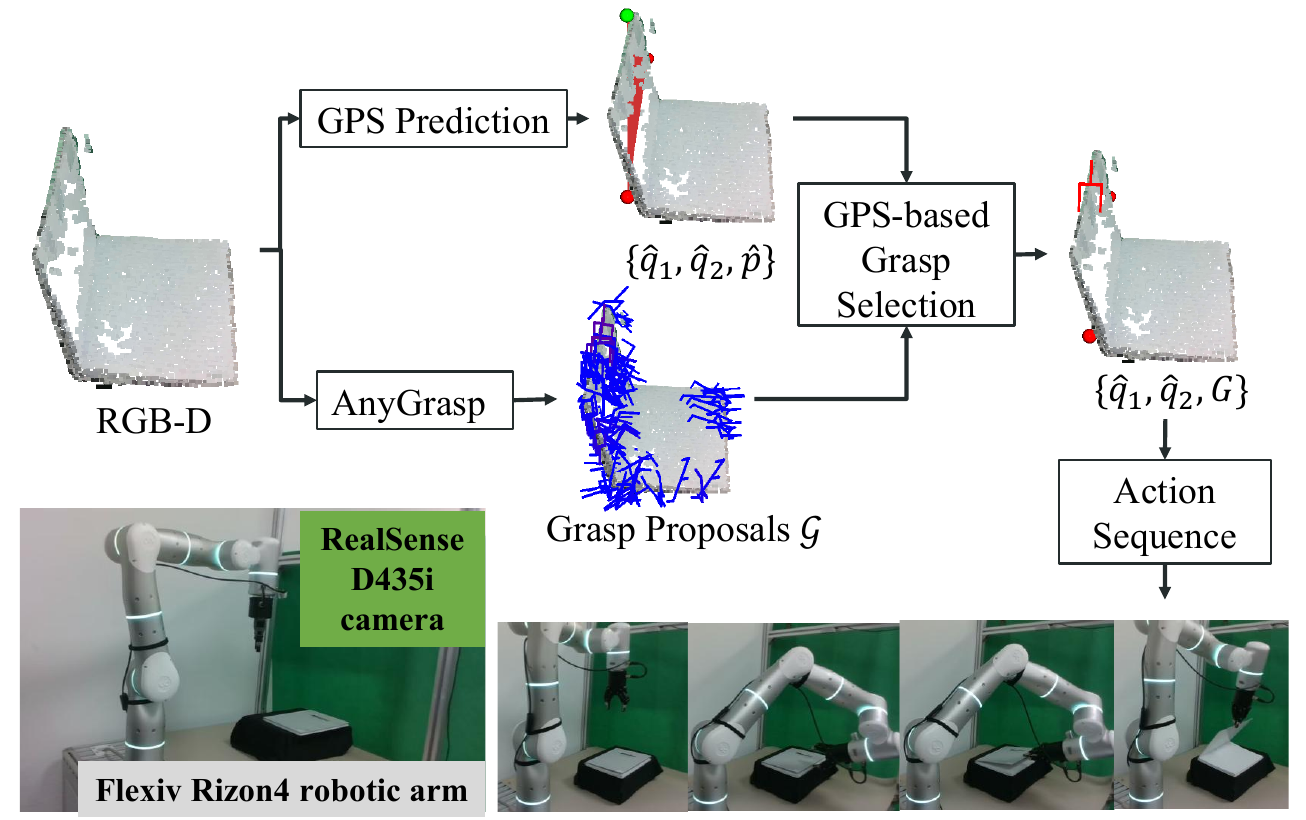}
\caption{Heuristic manipulation policy based on GPS prediction.}
\label{fig:robot_manip}
\end{center}
\vspace{-15pt}
\end{figure}

\begin{figure*}[t]
\begin{center}
\includegraphics[width=0.99\linewidth]{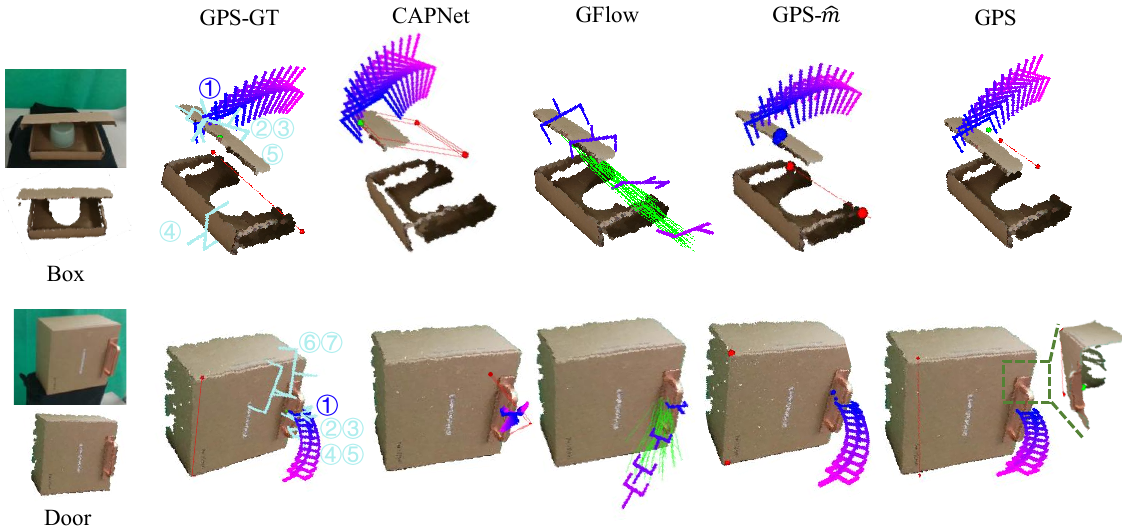}
\caption{
\textbf{Visualization results of robot experiments.}
Waypoints $\mathcal{T}_{\mathbf{G}} = \{\mathbf{T}_t\}_{t=1}^{\mathbf{t}}$ are depicted in gradient color from blue to purple.
In GPS-GT and GPS, red points denote ${\hat{\mathbf{q}}_1, \hat{\mathbf{q}}_2}$, a green point denotes $\hat{\mathbf{p}}$, and cyan grasps marks different initial grasp poses. For Door task in GPS, an additional view is provided to clearly display the otherwise occluded $\hat{\mathbf{p}}$.
In CAPNet, the predicted part bounding boxes are shown in red.
In GFlow, the predicted flow is visualized as a green line, spanning a total of four steps.
In GPS-$\hat{\mathbf{m}}$, the blue points indicate  $\hat{\mathbf{m}}$.
}
\label{fig:robo_vis}
\end{center}
\vspace{-10pt}
\end{figure*}

\subsection{Heuristic Policy}
\label{sec:heu_policy} 
For a robot to manipulate an object in the real physical world, 
we plan feasible robot trajectories based on predicted GPS $\{\hat{\mathbf{q}_1}, \hat{\mathbf{q}_2}, \hat{\mathbf{p}}\}$. The process is detailed in Fig.~\ref{fig:robot_manip}.
This involved two modules: 
\textbf{1) Initial grasp selection} to decide where to grasp the objects.
We use AnyGrasp~\cite{fang2023anygrasp} to generate grasp proposals $\mathcal{G}=\{\mathbf{G}_{k}\}_{k=1}^{K}$.
Then, GPS predictions are used to select the best initial grasp $\mathbf{G}$, which corresponds to an end-effector pose $\mathbf{T}_1$.
To select $\mathbf{G}$, GPS predictions are used for a scoring function of grasp proposals. 
A good grasp should be close to the plane defined by $\{\hat{\mathbf{q}_1}, \hat{\mathbf{q}_2}, \hat{\mathbf{p}}\}$.
The additional constraints are combined with the original grasp confidence scores to form the final grasp scores.
For objects with a prismatic joint, the criterion is the distance to the $\{\hat{\mathbf{q}_1}, \hat{\mathbf{q}_2}, \hat{\mathbf{p}}\}$ plane
; 
\textbf{2) Waypoints generation} to decide how to move objects after grasping.
After grasping the target part, the manipulation policy explicitly utilizes $\{\hat{\mathbf{q}_1}, \hat{\mathbf{q}_2}\}$ and the robot's current state to generate action sequences $\mathcal{T}_{\mathbf{G}} = \{\mathbf{T}_t\}_{t=1}^{\mathbf{t}}$  based on Eq.~\ref{eq:revolute_motion}, ~\ref{eq:prism_motion}.

\subsection{Settings}
For real-robot experiments, we set up one Flexiv Rizon4 arm equipped with a gripper and an Intel RealSense D435 RGB-D camera. As shown in Fig.~\ref{fig:robot_manip}, the camera is mounted on the wrist of the robotic arm, which was calibrated in an eye-in-hand configuration. 
GPS and AnyGrasp~\cite{fang2023anygrasp} prediction results are in the camera frame; thus, we transform them from the camera frame into the robot base frame using the calibrated hand-eye matrix.  

We test on 9 objects with diverse appearances. Their categories and part classes are:
Box (Lid), Document-Box (Lid), Bucket (Handle), Door (Door), Drawer (Drawer), Notebook (Lid-book), Folder (Lid-book), Lamp (Lid-thin), Clapperboard (Lid-thin). 
An object is successfully manipulated if its part is rotated by 50\degree (revolute axis) or moved 5cm (prismatic axis).
Each waypoint rotates a part by 5\degree  (revolute axis) or move it by 0.5cm (prismatic axis).
The camera is moved to different views to obtain the object point cloud before manipulation.
Each object is tested with 30 trials, each trial different state, \ie, a  combination of 6 different camera views and 5 different initial part poses. 

After generating the waypoints $\mathcal{T}_{\mathbf{G}} = \{\mathbf{T}_t\}_{t=1}^{\mathbf{t}}$, we check it with planning algorithm RRT*\cite{rrt_star} to avoid kinematically infeasible trajectory. 
If verified, the trajectory is executed. Otherwise, we directly mark this trial as a failure. 

\begin{table*}[t]
    \centering
     \resizebox{0.99\linewidth}{!}{
      \begin{tabular}{l | ccccc | cccc | cc}
        \toprule
        \textbf{Method}  & \textbf{Box} & \textbf{Document-Box} & \textbf{Bucket} & \textbf{Door} & \textbf{Drawer} & \textbf{Notebook} & \textbf{Folder} & \textbf{Lamp} & \textbf{Clapperboard} & \textbf{Avg-overlap} & \textbf{Avg-all}\\
        \midrule
         GPS-GT   & 93\% &	93\% &	87\% &	93\% &	87\% &	90\% &	90\% &	93\% &	97\% & 91\% & 91\% 
         \\
         \midrule
         CAPNet~\cite{capnet}  & 60\% &	30\% &	63\% &	40\% &	43\% &	13\% &	27\% &	13\% &	7\% & 47\% & 33\% 
         \\
         GFlow~\cite{yuan2024general}  & 60\% &	37\% &	60\% &	17\% &	53\% &	17\% &	27\% &	23\% &	20\% &	45\% & 35\%
         \\
         GPS-${\hat{\mathbf{m}}}$  &  77\% &	57\% &	53\% &	53\% &	50\% &	63\% &	70\% &	33\% &	67\% &	58\% & 58\% 
         \\
         GPS   &  \textbf{93\%} & \textbf{90\%} & \textbf{67\%} & \textbf{67\%} & \textbf{60\%} & \textbf{73\%} & \textbf{70\%} & \textbf{67\%} & \textbf{73\%} & \textbf{75\%} & \textbf{73\%}\\
        \bottomrule
      \end{tabular}}
      \caption{\textbf{Robot manipulation successful rate.} We test on 9 objects, each with 30 trials. The first 5 objects are overlapped categories with baselines~\cite{capnet,yuan2024general}. We average the success rate on them as ``Avg-overlap''. The successful rate for all the 9 objects is ``Avg-all''.
      }
  \label{tab:robot}
\end{table*}

\subsection{Results}
We use results from different perception methods to generate initial grasp and axis-guided waypoints, and use the success rate to measure perception model performance.  The results and visualization are shown in Tab~\ref{tab:robot} and Fig.~\ref{fig:robo_vis}.
We mainly answer three questions:
\textbf{1)} Can we find a good initial grasp from the loose plane constraint from GPS?
\textbf{2)} How does GPS perform as a perception module in robot manipulation?
\textbf{3)} What are the typical failure cases?

\subsubsection{Initial Grasp Evaluation}
To verify GPS's ability to find a good initial grasp with a loose plane constraint, we use human-annotated GPS-GT to generate initial grasp and waypoints, execute, and report the  success rate. 
As GPS-GT is not predicted by a GPS model, we can exclude GPS prediction error and verify the GPS representation itself.
Two visualization results for the box and door are illustrated in Fig.~\ref{fig:robo_vis}.
The box example demonstrates how the $\mathbf{p}$-constraint effectively modulates grasp selection. While the cyan grasp $G_4$ has the highest original confidence, it drops to the 4th rank after the GPS-based geometric re-scoring, leading to the selection of the top-ranked blue grasp $G_1$ instead.
The door example shows the importance of the original grasp of confidence. Grasps $G_6$ and $G_7$, though geometrically close to $\mathbf{p}$, are assigned low final scores due to their low grasp confidence.
Tab.~\ref{tab:robot} reports an overall 91\% success rate over 270 trials, verifying the effectiveness of GPS even in a geometrically loose form.

\subsubsection{Performance Comparison}
We next evaluate our method by replacing GPS-GT with predictions from a learned \textbf{GPS} model. As is shown in Tab.~\ref{tab:robot}, our approach achieves an average success rate of 73\% without any in-domain fine-tuning. Fig.~\ref{fig:robo_vis} shows that the predicted GPS closely aligns with the ground-truth annotation, leading to similar successful trajectories.
Below we compare against three baselines: 
CAPNet~\cite{capnet}, GFlow~\cite{yuan2024general} and GPS-$\hat{\mathbf{m}}$.
They are all related to select grasp from a given or inferred contact point $\mathbf{m}$. 
We calculate the distance between a grasp and the contact point, and combine it with the original grasp score to select $\mathbf{G}$.

\textbf{CAPNet}~\cite{capnet} predicts the part bounding box, from which we derive the articulation axis and contact point.
For the door example Fig.~\ref{fig:robo_vis}, CAPNet fails to correctly recognize the closed door structure, leading to an invalid prediction. 
For the box example, the estimated axis is inaccurate, causing the robot to bend the lid rather than open it properly. 
Overall, CAPNet attains only a 33\% success rate across the tested objects, primarily  due to the sim-to-real gap.

For \textbf{GFlow}~\cite{yuan2024general},
we use the heuristic policy in its original paper: a contact point $\mathbf{m}$ is manually given, query points near $\mathbf{m}$ are selected to predict scene flow, and Singular Value Decomposition is used to align the end-effector’s motion with the predicted flow, which has 4 execution steps.
Despite the given contact point (actually unfair for comparison), GFlow~\cite{yuan2024general} is still prone to errors, with a 
35\% 
success rate. This is largely due to its training on limited data diversity and the inherent difficulty in learning reliable flow fields. 
As is shown in Fig.~\ref{fig:robo_vis}, the predicted flow for the box deviate significantly, and the door flow deviates downward.

\begin{figure}[t]
\begin{center}
\includegraphics[width=0.95\linewidth]{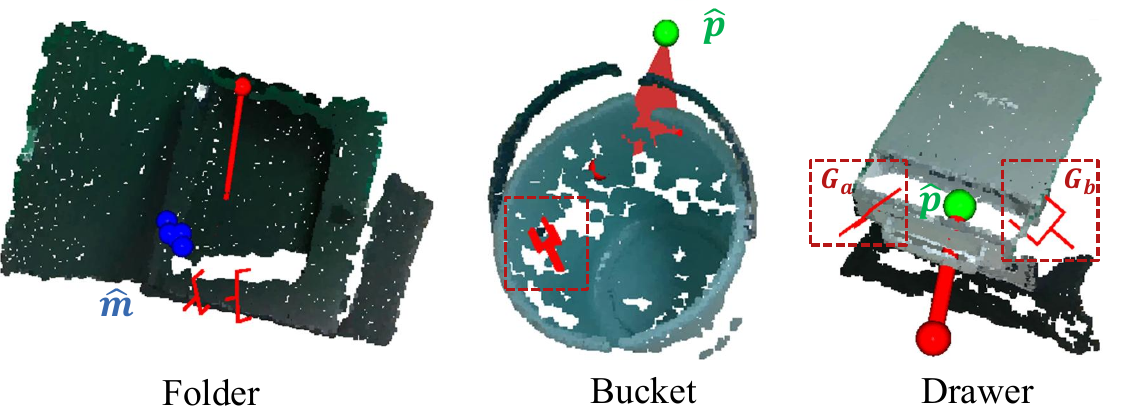}
\caption{
\textbf{Failure cases.}
The folder is a failure example for GPS-$\hat{\mathbf{m}}$, and the bucket and drawer are failure examples for GPS.
Red points are $\{\hat{\mathbf{q}}_1, \hat{\mathbf{q}}_2\}$, green points are $\hat{\mathbf{p}}$, and blue points are $\hat{\mathbf{m}}$.
}
\label{fig:robo_failure}
\end{center}
\vspace{-15pt}
\end{figure}

For \textbf{GPS-$\hat{\mathbf{m}}$}, we conduct ablation study to predict $\mathbf{m}$ instead of predicting $\mathbf{p}$, as mentioned in Sec.~\ref{sec:revolute}. 
To acquire ground-truth $\mathbf{m}$, we fit a Gaussian mixture model (GMM)~\cite{bahl2023affordances} to the closest 200 points around $\mathbf{p}$, parameterized by $\{\mu_i, \Sigma_i\}_{i=1}^5$. The model predicts $\{\mu_i\}_{i=1}^5$ coordinates instead of $\mathbf{p}$.
The learned model  performs poorer with a 58\% success rate.
The folder example in Fig.~\ref{fig:robo_failure} is a typical failure case, where the prediction $\hat{\mathbf{m}}$ is wrong and far from easily graspable edges.
Thus, using loose geometric constraints $\mathbf{p}$ is more generalized for manipulation.


\subsubsection{Failure Case Analysis}
We show failure cases in Fig.~\ref{fig:robo_failure}. 
GPS is not predicted well under a few part poses and camera views, mainly influenced by point cloud noise, \eg, the bucket example.
In other cases, GPS fails to select a proper grasp despite a correct prediction. 
For example, $\hat{\mathbf{p}}$-error is small for the drawer, but a wrong grasp $G_b$ near the body is selected, instead of a more reasonable grasp $G_a$.
This is because the scoring function is not flexible enough to balance between grasp score and geometric constraints,
which can be improved by fine-tuning the grasping model with GPS input,  or integrating GPS with more advanced methods (\eg, diffusion policy~\cite{dp,wang2024rise}, VLA model~\cite{kim2024openvla}) in future work.

\section{Conclusion}
This paper proposes a novel affordance representation GPS for articulated part estimation. It balances data scalability with annotation quality.
With a data-efficient system integrated with VR device, we collect the VR-GPS dataset with rich object knowledge. 
The learned GPS model has better perception performance and can facilitate a robot to manipulate daily objects via a heuristic policy.

\section{Acknowledgments}
We sincerely thank Ziyu Wang, Hongjie Fang, Xinyu Zhan, Shizheng Zhu for their help in VR-GPS system construction and verification. 

This work was supported in part by Fundamental and Interdisciplinary Disciplines Breakthrough Plan of the Ministry of Education of China, 
the National Natural Science Foundation of China under Grant No.~U25A20442, 62306175,
Shanghai Municipal Science and Technology Major Project No.~2025SHZDZX025G14.

{
    \small
    \bibliographystyle{ieeenat_fullname}
    \bibliography{main}

@inproceedings{liu2022hoi4d,
  title={Hoi4d: A 4d egocentric dataset for category-level human-object interaction},
  author={Liu, Yunze and Liu, Yun and Jiang, Che and Lyu, Kangbo and Wan, Weikang and Shen, Hao and Liang, Boqiang and Fu, Zhoujie and Wang, He and Yi, Li},
  booktitle={Proceedings of the IEEE/CVF Conference on Computer Vision and Pattern Recognition},
  pages={21013--21022},
  year={2022}
}

@inproceedings{zhan2024oakink2,
  title={Oakink2: A dataset of bimanual hands-object manipulation in complex task completion},
  author={Zhan, Xinyu and Yang, Lixin and Zhao, Yifei and Mao, Kangrui and Xu, Hanlin and Lin, Zenan and Li, Kailin and Lu, Cewu},
  booktitle={Proceedings of the IEEE/CVF Conference on Computer Vision and Pattern Recognition},
  pages={445--456},
  year={2024}
}

@inproceedings{liu2024taco,
  title={Taco: Benchmarking generalizable bimanual tool-action-object understanding},
  author={Liu, Yun and Yang, Haolin and Si, Xu and Liu, Ling and Li, Zipeng and Zhang, Yuxiang and Liu, Yebin and Yi, Li},
  booktitle={Proceedings of the IEEE/CVF Conference on Computer Vision and Pattern Recognition},
  pages={21740--21751},
  year={2024}
}

@article{wen2023any,
  title={Any-point trajectory modeling for policy learning},
  author={Wen, Chuan and Lin, Xingyu and So, John and Chen, Kai and Dou, Qi and Gao, Yang and Abbeel, Pieter},
  journal={arXiv preprint arXiv:2401.00025},
  year={2023}
}

@inproceedings{geng2023gapartnet,
  title={Gapartnet: Cross-category domain-generalizable object perception and manipulation via generalizable and actionable parts},
  author={Geng, Haoran and Xu, Helin and Zhao, Chengyang and Xu, Chao and Yi, Li and Huang, Siyuan and Wang, He},
  booktitle={Proceedings of the IEEE/CVF Conference on Computer Vision and Pattern Recognition},
  pages={7081--7091},
  year={2023}
}

@inproceedings{xiang2020sapien,
  title={Sapien: A simulated part-based interactive environment},
  author={Xiang, Fanbo and Qin, Yuzhe and Mo, Kaichun and Xia, Yikuan and Zhu, Hao and Liu, Fangchen and Liu, Minghua and Jiang, Hanxiao and Yuan, Yifu and Wang, He and others},
  booktitle={Proceedings of the IEEE/CVF conference on computer vision and pattern recognition},
  pages={11097--11107},
  year={2020}
}

@article{qi2017pointnet++,
  title={Pointnet++: Deep hierarchical feature learning on point sets in a metric space},
  author={Qi, Charles Ruizhongtai and Yi, Li and Su, Hao and Guibas, Leonidas J},
  journal={Advances in neural information processing systems},
  volume={30},
  year={2017}
}

@article{yuan2024general,
  title={General flow as foundation affordance for scalable robot learning},
  author={Yuan, Chengbo and Wen, Chuan and Zhang, Tong and Gao, Yang},
  journal={arXiv preprint arXiv:2401.11439},
  year={2024}
}

@inproceedings{bahl2023affordances,
  title={Affordances from human videos as a versatile representation for robotics},
  author={Bahl, Shikhar and Mendonca, Russell and Chen, Lili and Jain, Unnat and Pathak, Deepak},
  booktitle={Proceedings of the IEEE/CVF Conference on Computer Vision and Pattern Recognition},
  pages={13778--13790},
  year={2023}
}

@article{wang2024shape,
  title={Shape of motion: 4d reconstruction from a single video},
  author={Wang, Qianqian and Ye, Vickie and Gao, Hang and Austin, Jake and Li, Zhengqi and Kanazawa, Angjoo},
  journal={arXiv preprint arXiv:2407.13764},
  year={2024}
}

@article{fang2023anygrasp,
  title={Anygrasp: Robust and efficient grasp perception in spatial and temporal domains},
  author={Fang, Hao-Shu and Wang, Chenxi and Fang, Hongjie and Gou, Minghao and Liu, Jirong and Yan, Hengxu and Liu, Wenhai and Xie, Yichen and Lu, Cewu},
  journal={IEEE Transactions on Robotics},
  volume={39},
  number={5},
  pages={3929--3945},
  year={2023},
  publisher={IEEE}
}

@article{chen2024arcap,
  title={Arcap: Collecting high-quality human demonstrations for robot learning with augmented reality feedback},
  author={Chen, Sirui and Wang, Chen and Nguyen, Kaden and Fei-Fei, Li and Liu, C Karen},
  journal={arXiv preprint arXiv:2410.08464},
  year={2024}
}

@article{ravi2024sam2,
  title={SAM 2: Segment Anything in Images and Videos},
  author={Ravi, Nikhila and Gabeur, Valentin and Hu, Yuan-Ting and Hu, Ronghang and Ryali, Chaitanya and Ma, Tengyu and Khedr, Haitham and R{\"a}dle, Roman and Rolland, Chloe and Gustafson, Laura and Mintun, Eric and Pan, Junting and Alwala, Kalyan Vasudev and Carion, Nicolas and Wu, Chao-Yuan and Girshick, Ross and Doll{\'a}r, Piotr and Feichtenhofer, Christoph},
  journal={arXiv preprint arXiv:2408.00714},
  url={https://arxiv.org/abs/2408.00714},
  year={2024}
}

@inproceedings{clip,
  title={Learning transferable visual models from natural language supervision},
  author={Radford, Alec and Kim, Jong Wook and Hallacy, Chris and Ramesh, Aditya and Goh, Gabriel and Agarwal, Sandhini and Sastry, Girish and Askell, Amanda and Mishkin, Pamela and Clark, Jack and others},
  booktitle={International conference on machine learning},
  pages={8748--8763},
  year={2021},
  organization={PmLR}
}

@article{rsrd,
  title={Robot see robot do: Imitating articulated object manipulation with monocular 4d reconstruction},
  author={Kerr, Justin and Kim, Chung Min and Wu, Mingxuan and Yi, Brent and Wang, Qianqian and Goldberg, Ken and Kanazawa, Angjoo},
  journal={arXiv preprint arXiv:2409.18121},
  year={2024}
}

@inproceedings{fang2024rh20t,
  title={Rh20t: A comprehensive robotic dataset for learning diverse skills in one-shot},
  author={Fang, Hao-Shu and Fang, Hongjie and Tang, Zhenyu and Liu, Jirong and Wang, Chenxi and Wang, Junbo and Zhu, Haoyi and Lu, Cewu},
  booktitle={2024 IEEE International Conference on Robotics and Automation (ICRA)},
  pages={653--660},
  year={2024},
  organization={IEEE}
}

@article{dp,
  title={Diffusion policy: Visuomotor policy learning via action diffusion},
  author={Chi, Cheng and Xu, Zhenjia and Feng, Siyuan and Cousineau, Eric and Du, Yilun and Burchfiel, Benjamin and Tedrake, Russ and Song, Shuran},
  journal={The International Journal of Robotics Research},
  pages={02783649241273668},
  year={2023},
  publisher={SAGE Publications Sage UK: London, England}
}

@article{kim2024openvla,
  title={Openvla: An open-source vision-language-action model},
  author={Kim, Moo Jin and Pertsch, Karl and Karamcheti, Siddharth and Xiao, Ted and Balakrishna, Ashwin and Nair, Suraj and Rafailov, Rafael and Foster, Ethan and Lam, Grace and Sanketi, Pannag and others},
  journal={arXiv preprint arXiv:2406.09246},
  year={2024}
}

@article{makoviychuk2021isaac,
  title={Isaac gym: High performance gpu-based physics simulation for robot learning},
  author={Makoviychuk, Viktor and Wawrzyniak, Lukasz and Guo, Yunrong and Lu, Michelle and Storey, Kier and Macklin, Miles and Hoeller, David and Rudin, Nikita and Allshire, Arthur and Handa, Ankur and others},
  journal={arXiv preprint arXiv:2108.10470},
  year={2021}
}

@inproceedings{shen2021igibson,
  title={igibson 1.0: A simulation environment for interactive tasks in large realistic scenes},
  author={Shen, Bokui and Xia, Fei and Li, Chengshu and Mart{\'\i}n-Mart{\'\i}n, Roberto and Fan, Linxi and Wang, Guanzhi and P{\'e}rez-D’Arpino, Claudia and Buch, Shyamal and Srivastava, Sanjana and Tchapmi, Lyne and others},
  booktitle={2021 IEEE/RSJ International Conference on Intelligent Robots and Systems (IROS)},
  pages={7520--7527},
  year={2021},
  organization={IEEE}
}

@article{zhang2024monst3r,
  title={Monst3r: A simple approach for estimating geometry in the presence of motion},
  author={Zhang, Junyi and Herrmann, Charles and Hur, Junhwa and Jampani, Varun and Darrell, Trevor and Cole, Forrester and Sun, Deqing and Yang, Ming-Hsuan},
  journal={arXiv preprint arXiv:2410.03825},
  year={2024}
}

@inproceedings{capnet,
  title={CAP-Net: A Unified Network for 6D Pose and Size Estimation of Categorical Articulated Parts from a Single RGB-D Image},
  author={Huang, Jingshun and Lin, Haitao and Wang, Tianyu and Fu, Yanwei and Xue, Xiangyang and Zhu, Yi},
  booktitle={Proceedings of the Computer Vision and Pattern Recognition Conference},
  pages={11654--11664},
  year={2025}
}

@article{rrt_star,
  title={Sampling-based algorithms for optimal motion planning},
  author={Karaman, Sertac and Frazzoli, Emilio},
  journal={The international journal of robotics research},
  volume={30},
  number={7},
  pages={846--894},
  year={2011},
  publisher={Sage Publications Sage UK: London, England}
}

@article{featup,
  title={Featup: A model-agnostic framework for features at any resolution},
  author={Fu, Stephanie and Hamilton, Mark and Brandt, Laura and Feldman, Axel and Zhang, Zhoutong and Freeman, William T},
  journal={arXiv preprint arXiv:2403.10516},
  year={2024}
}

@article{trace-anything,
  title={Trace Anything: Representing Any Video in 4D via Trajectory Fields},
  author={Liu, Xinhang and Xiao, Yuxi and Chen, Donny Y and Feng, Jiashi and Tai, Yu-Wing and Tang, Chi-Keung and Kang, Bingyi},
  journal={arXiv preprint arXiv:2510.13802},
  year={2025}
}

@article{onepose++,
  title={Onepose++: Keypoint-free one-shot object pose estimation without CAD models},
  author={He, Xingyi and Sun, Jiaming and Wang, Yuang and Huang, Di and Bao, Hujun and Zhou, Xiaowei},
  journal={Advances in Neural Information Processing Systems},
  volume={35},
  pages={35103--35115},
  year={2022}
}

@inproceedings{foundationpose,
  title={Foundationpose: Unified 6d pose estimation and tracking of novel objects},
  author={Wen, Bowen and Yang, Wei and Kautz, Jan and Birchfield, Stan},
  booktitle={Proceedings of the IEEE/CVF Conference on Computer Vision and Pattern Recognition},
  pages={17868--17879},
  year={2024}
}

@inproceedings{npcs,
  title={Category-level articulated object pose estimation},
  author={Li, Xiaolong and Wang, He and Yi, Li and Guibas, Leonidas J and Abbott, A Lynn and Song, Shuran},
  booktitle={Proceedings of the IEEE/CVF conference on computer vision and pattern recognition},
  pages={3706--3715},
  year={2020}
}

@inproceedings{mo2019partnet,
  title={Partnet: A large-scale benchmark for fine-grained and hierarchical part-level 3d object understanding},
  author={Mo, Kaichun and Zhu, Shilin and Chang, Angel X and Yi, Li and Tripathi, Subarna and Guibas, Leonidas J and Su, Hao},
  booktitle={Proceedings of the IEEE/CVF conference on computer vision and pattern recognition},
  pages={909--918},
  year={2019}
}

@inproceedings{akb48,
  title={Akb-48: A real-world articulated object knowledge base},
  author={Liu, Liu and Xu, Wenqiang and Fu, Haoyuan and Qian, Sucheng and Yu, Qiaojun and Han, Yang and Lu, Cewu},
  booktitle={Proceedings of the IEEE/CVF Conference on Computer Vision and Pattern Recognition},
  pages={14809--14818},
  year={2022}
}

@inproceedings{paris,
  title={Paris: Part-level reconstruction and motion analysis for articulated objects},
  author={Liu, Jiayi and Mahdavi-Amiri, Ali and Savva, Manolis},
  booktitle={Proceedings of the IEEE/CVF International Conference on Computer Vision},
  pages={352--363},
  year={2023}
}

@article{artgs,
  title={Artgs: Building interactable replicas of complex articulated objects via gaussian splatting},
  author={Liu, Yu and Jia, Baoxiong and Lu, Ruijie and Ni, Junfeng and Zhu, Song-Chun and Huang, Siyuan},
  journal={arXiv preprint arXiv:2502.19459},
  year={2025}
}

@inproceedings{ditto,
  title={Ditto: Building digital twins of articulated objects from interaction},
  author={Jiang, Zhenyu and Hsu, Cheng-Chun and Zhu, Yuke},
  booktitle={Proceedings of the IEEE/CVF Conference on Computer Vision and Pattern Recognition},
  pages={5616--5626},
  year={2022}
}

@inproceedings{nocs,
  title={Normalized object coordinate space for category-level 6d object pose and size estimation},
  author={Wang, He and Sridhar, Srinath and Huang, Jingwei and Valentin, Julien and Song, Shuran and Guibas, Leonidas J},
  booktitle={Proceedings of the IEEE/CVF conference on computer vision and pattern recognition},
  pages={2642--2651},
  year={2019}
}

@article{martin2019rbo,
  title={The RBO dataset of articulated objects and interactions},
  author={Mart{\'\i}n-Mart{\'\i}n, Roberto and Eppner, Clemens and Brock, Oliver},
  journal={The International Journal of Robotics Research},
  volume={38},
  number={9},
  pages={1013--1019},
  year={2019},
  publisher={SAGE Publications Sage UK: London, England}
}

@inproceedings{hinterstoisser2011multimodal,
  title={Multimodal templates for real-time detection of texture-less objects in heavily cluttered scenes},
  author={Hinterstoisser, Stefan and Holzer, Stefan and Cagniart, Cedric and Ilic, Slobodan and Konolige, Kurt and Navab, Nassir and Lepetit, Vincent},
  booktitle={2011 international conference on computer vision},
  pages={858--865},
  year={2011},
  organization={IEEE}
}

@inproceedings{calli2015ycb,
  title={The ycb object and model set: Towards common benchmarks for manipulation research},
  author={Calli, Berk and Singh, Arjun and Walsman, Aaron and Srinivasa, Siddhartha and Abbeel, Pieter and Dollar, Aaron M},
  booktitle={2015 international conference on advanced robotics (ICAR)},
  pages={510--517},
  year={2015},
  organization={IEEE}
}

@inproceedings{affordancenet,
  title={3d affordancenet: A benchmark for visual object affordance understanding},
  author={Deng, Shengheng and Xu, Xun and Wu, Chaozheng and Chen, Ke and Jia, Kui},
  booktitle={proceedings of the IEEE/CVF conference on computer vision and pattern recognition},
  pages={1778--1787},
  year={2021}
}

@inproceedings{ju2024robo,
  title={Robo-abc: Affordance generalization beyond categories via semantic correspondence for robot manipulation},
  author={Ju, Yuanchen and Hu, Kaizhe and Zhang, Guowei and Zhang, Gu and Jiang, Mingrun and Xu, Huazhe},
  booktitle={European Conference on Computer Vision},
  pages={222--239},
  year={2024},
  organization={Springer}
}

@article{xiao2025spatialtrackerv2,
  title={Spatialtrackerv2: 3d point tracking made easy},
  author={Xiao, Yuxi and Wang, Jianyuan and Xue, Nan and Karaev, Nikita and Makarov, Yuri and Kang, Bingyi and Zhu, Xing and Bao, Hujun and Shen, Yujun and Zhou, Xiaowei},
  journal={arXiv preprint arXiv:2507.12462},
  year={2025}
}

@article{ransac,
  title={Random sample consensus: a paradigm for model fitting with applications to image analysis and automated cartography},
  author={Fischler, Martin A and Bolles, Robert C},
  journal={Communications of the ACM},
  volume={24},
  number={6},
  pages={381--395},
  year={1981},
  publisher={ACM New York, NY, USA}
}

@article{umeyama,
  title={Least-squares estimation of transformation parameters between two point patterns},
  author={Umeyama, Shinji},
  journal={IEEE Transactions on pattern analysis and machine intelligence},
  volume={13},
  number={4},
  pages={376--380},
  year={2002},
  publisher={IEEE}
}

@inproceedings{wang2024rise,
  title={Rise: 3d perception makes real-world robot imitation simple and effective},
  author={Wang, Chenxi and Fang, Hongjie and Fang, Hao-Shu and Lu, Cewu},
  booktitle={2024 IEEE/RSJ International Conference on Intelligent Robots and Systems (IROS)},
  pages={2870--2877},
  year={2024},
  organization={IEEE}
}

@article{wu2023symbol,
  title={Symbol-LLM: leverage language models for symbolic system in visual human activity reasoning},
  author={Wu, Xiaoqian and Li, Yong-Lu and Sun, Jianhua and Lu, Cewu},
  journal={Advances in neural information processing systems},
  volume={36},
  pages={29680--29691},
  year={2023}
}

@inproceedings{grauman2022ego4d,
  title={Ego4d: Around the world in 3,000 hours of egocentric video},
  author={Grauman, Kristen and Westbury, Andrew and Byrne, Eugene and Chavis, Zachary and Furnari, Antonino and Girdhar, Rohit and Hamburger, Jackson and Jiang, Hao and Liu, Miao and Liu, Xingyu and others},
  booktitle={Proceedings of the IEEE/CVF conference on computer vision and pattern recognition},
  pages={18995--19012},
  year={2022}
}

@inproceedings{shan2020understanding,
  title={Understanding human hands in contact at internet scale},
  author={Shan, Dandan and Geng, Jiaqi and Shu, Michelle and Fouhey, David F},
  booktitle={Proceedings of the IEEE/CVF conference on computer vision and pattern recognition},
  pages={9869--9878},
  year={2020}
}
}


\maketitlesupplementary


\section{Detailed Comparison with Existing Works}



\begin{figure}[t]
\begin{center}
\includegraphics[width=0.8\columnwidth]{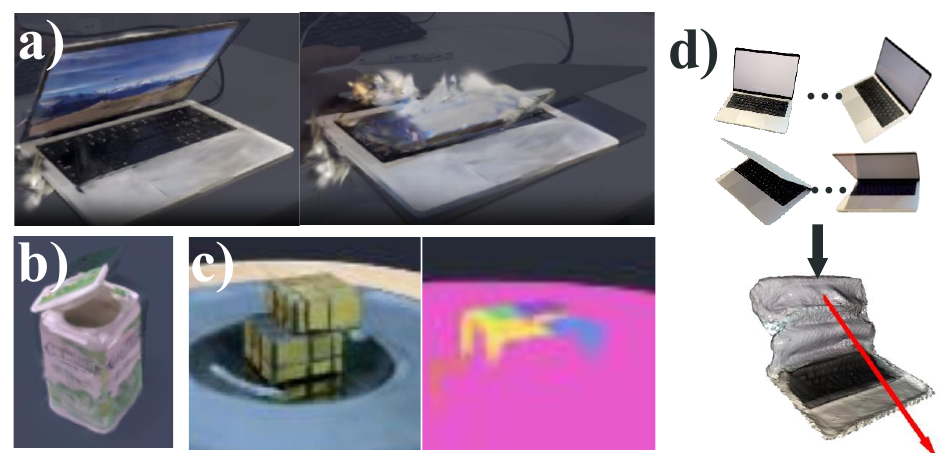}
\caption{Failure cases for post-processing methods RSRD~\cite{rsrd} and ArtGS~\cite{artgs}, which are prone to errors.}
\label{fig:rsrd}
\end{center}
\end{figure}

For pose-based representation, post-processing methods have emerged to reconstruct articulated objects from visual inputs. However, our method has unique advantages.

RSRD~\cite{rsrd} uses a 4D differentiable part model to recover object motions from an object scan and a single monocular video. It is time-consuming. Reconstruction takes about 40 minutes on a single 3090 GPU following its official code, and pose estimation (10 minutes) is needed for each interaction sequence.
Each time we interact with the object in another environment or camera view, we need to re-run pose estimation (10 minutes).
Furthermore, this optimization-based method is prone to errors, \textit{e.g.}, Fig.~\ref{fig:rsrd} (a) self-occlusion; (b) initial frame error; (c) segmentation error.
Instead, our VR-GPS are quick and ensure quality with manual annotation.

There are other post-processing methods.
Ditto\cite{ditto} fails to process most of the VR-GPS objects, \textit{e.g.}, book, lamp, as it is trained only on 8 categories and needs to train a network for each category.
PARIS\cite{paris} and ArtGS\cite{artgs}
leverages neural radiance fields and 3D Gaussians to reconstruct objects and estimate joints.
They excel in synthetic objects, but fail to estimate joints in real-world scenes (Fig.~\ref{fig:rsrd}(d)) and take an extra 20 minutes to manually align two states.


\section{Detailed Dataset Statistics}
VR-GPS is developed in Unity and deployed on a Meta Quest 3 device, based on the existing work~\cite{chen2024arcap}.
The virtual point coordinate is in the world frame determined during each initial configuration. During interaction, the relative transformation of the world frame and the headset is recorded. With the fixed transformation of the headset and the RealSense camera, the virtual point coordinate can finally be mapped to the camera frame.
As an intermediate frame, the world frame can be located anywhere within the boundary. We also provide VR recording videos as an attachment in the supplementary material.

\begin{figure}[t]
\begin{center}
\includegraphics[width=1\columnwidth]{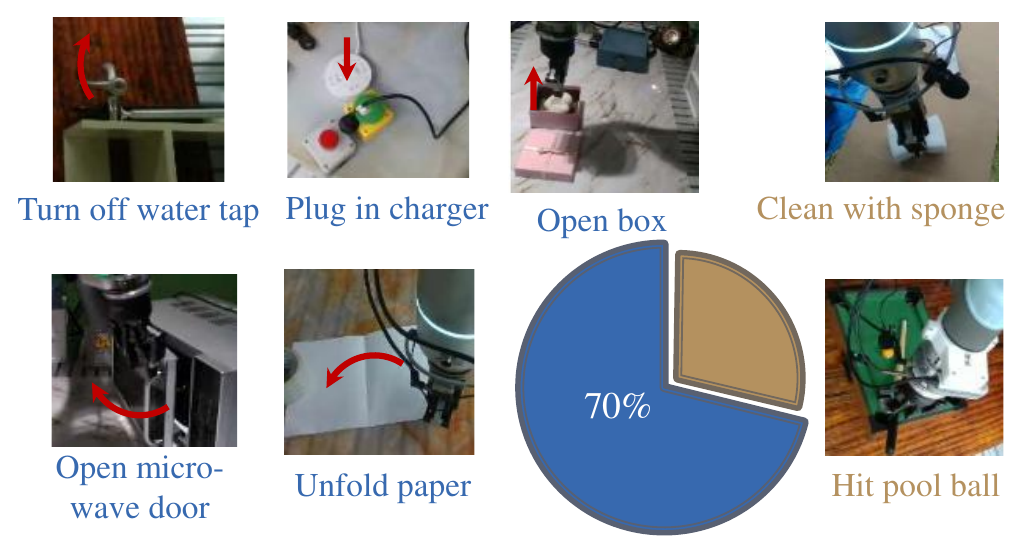}
\caption{A large proportion of current robot tasks are related to object geometric structure. Tasks with precise force control, \eg, hit a pool ball, are out of scope of this work.}
\label{fig:task_cover}
\end{center}
\end{figure}

To collect VR-GPS dataset, we invite 8 volunteers to annotate data wearing a headset, and another 3 volunteers to check.
The collected dataset has six part classes: Lid (89 objects), Lid-thin (21 objects), Lid-book (32 objects), Handle (34 objects), Door (33 objects), Drawer (25 objects).
 A large proportion of current robot tasks are related to the object's geometric structure. As is illustrated in Fig.~\ref{fig:task_cover}, among 89 complex tasks in RH20T~\cite{fang2024rh20t}, there are 70\% tasks requiring geometric structure knowledge, e.g., unfolding paper, plugging in a charger.

\section{Geometric Structure Learning}
\subsection{Benchmark Details}
We evaluate the model on two external datasets: HOI4D and RGBD-Art.
HOI4D has 1.2K frames for Laptop, 1.4K frames for Trashcan, 2.9K frames for Safe, 0.4K frames for Bucket, 2.8K frames for Drawer.   
RGBD-Art has 1.1K frames for Laptop, 0.6K frames for Trashcan, 0.5K frames for Safe, 1.4K frames for Bucket, 1.4K frames for Drawer.   

\subsection{Implementation Details}

We train our model on 2 NVIDIA H100 GPUs for a total of 100 epochs, using a batch size of 16.
The initial learning rate is set to 0.0001, using a warm-up
scheduler for gradual increase at the start of training. Input images are cropped and resized to 640$\times$640 resolution, and point clouds are randomly sampled to 24,576 points before being processed by the network.

\subsection{Transform Flow into GPS}
We transform flow prediction into GPS for comparison under the same metric. 
We first sample 1024 points on the object's surface using Farthest Point Sampling (FPS) and predict their trajectories.
To ensure quality and filter out static parts, we select $K=256$ trajectories with the largest total displacements. Then the GPS is extracted as follows:
For revolute objects, the rotation axis direction $\mathbf{u}$ is computed via Principal Component Analysis (PCA) on all motion vectors $\{\mathbf{d}_{j,t}\}$, corresponding to the eigenvector with the smallest eigenvalue:
\vspace{-5pt}
\begin{equation}
\label{eq:gflow_axis_dir_pca}
\mathbf{u} = \arg\min_{\|\mathbf{v}\|=1} \sum_{j,t} (\mathbf{v} \cdot \mathbf{d}_{j,t})^2,
\end{equation}
Second, we determine a point on the axis, $\mathbf{q}$, using a per-trajectory voting scheme. For each of the $K$ selected trajectories $\mathcal{T}_j$, we estimate an axis position candidate $\mathbf{c}_j$ by finding the common intersection of internal motion-perpendicular lines. The final axis point is the mean of these candidates:
\begin{equation}
\label{eq:gflow_axis_pos_vote}
\mathbf{q} = \frac{1}{K} \sum_{j=1}^{K} \mathbf{c}_j, \quad \text{where } \mathbf{c}_j = \mathcal{F}(\mathcal{T}_j, \mathbf{u}).
\end{equation}

For prismatic joints, the axis direction is instead the eigenvector with the largest eigenvalue, as motion is parallel to the axis. The axis position is the average of trajectory centers projected onto the perpendicular plane.

\begin{figure}[t]
\begin{center}
\includegraphics[width=0.99\columnwidth]{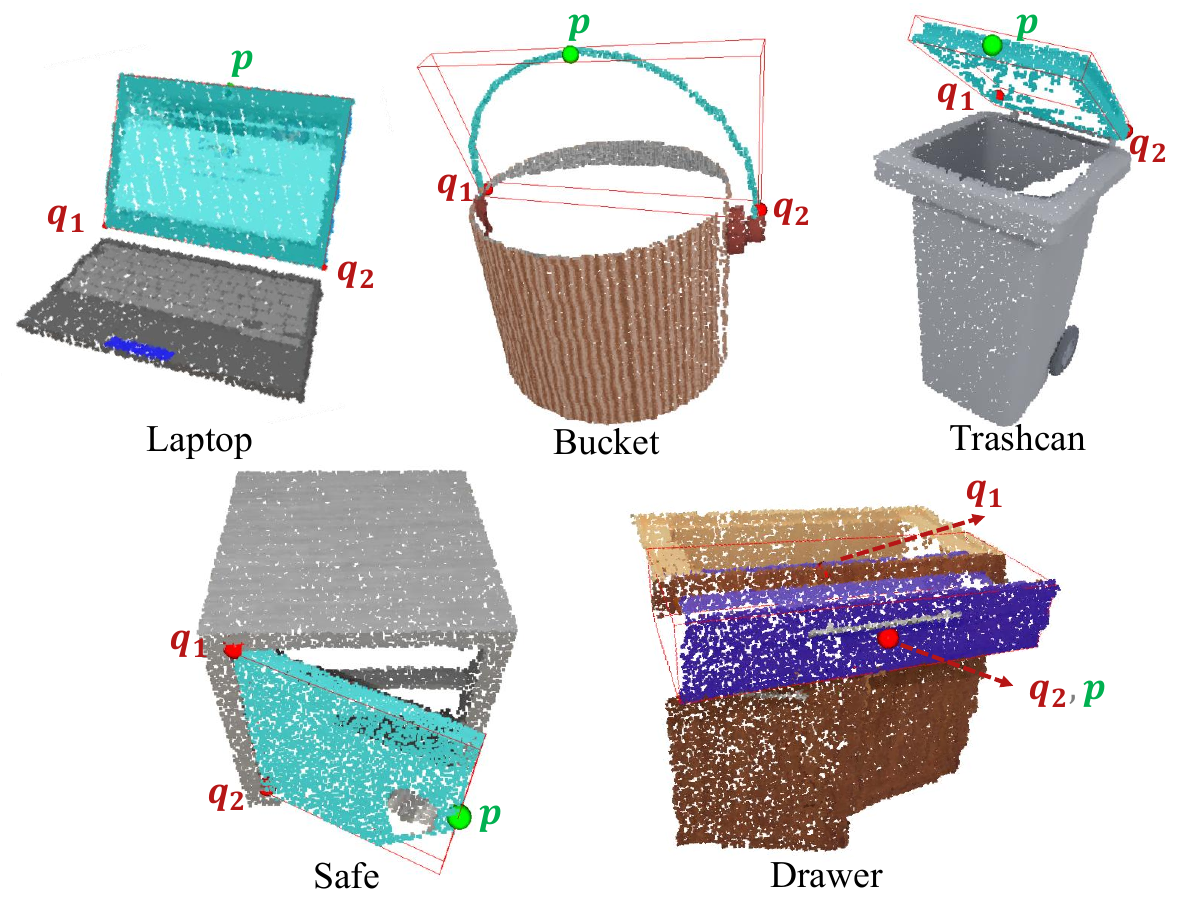}
\caption{Object part pose and corresponding GPS.}
\label{fig:transform_part_pose}
\end{center}
\end{figure}

\subsection{Transform Part Pose into GPS} 
We transform part pose into GPS for comparison under the same metric.
With the predicted part segmentation and NPCS~\cite{npcs} map, we apply RANSAC~\cite{ransac} for outlier removal and Umeyama algorithm~\cite{umeyama} to obtain part bounding box, and then calculate GPS from bounding box coordinates, as is shown in Fig.~\ref{fig:transform_part_pose}.

\section{Real Robot Experiments}

\subsection{Heuristic Policy}
\begin{figure}[t]
\begin{center}
\includegraphics[width=0.9\columnwidth]{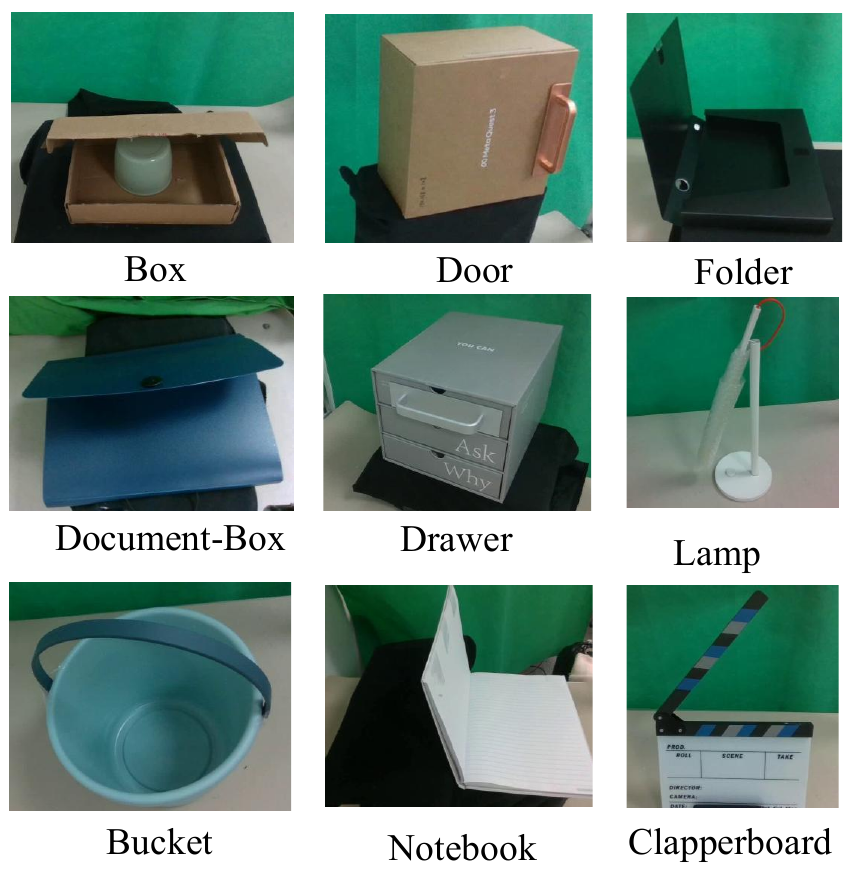}
\caption{The test objects in real robot experiments. We show a random view for each object.
}
\label{fig:robot_objects}
\end{center}
\end{figure}

\begin{algorithm}[t]
    \caption{Heuristic Policy}\label{alg:policy}
    \begin{algorithmic}[1]
        \REQUIRE{Object point cloud ${\mathcal{P} \in \mathbb{R} ^{N\times 3}}$; Time step $\mathbf{t}$;}
        \ENSURE{Planned robot trajectory}
        \STATE $\mathcal{G} \gets \texttt{AnyGrasp}(\mathcal{P})$
        \STATE $\{\hat{\mathbf{q}_1}, \hat{\mathbf{q}_2}, \hat{\mathbf{p}}\}$ $\gets$ $\texttt{GPS}(\mathcal{P})$
        \STATE $\mathbf{G}, \mathbf{T}_1 \gets \mathrm{arg}\max{\mathcal{S}_{\{\hat{\mathbf{q}_1}, \hat{\mathbf{q}_2}, \hat{\mathbf{p}}\}}(\mathcal{G})}$
        \FOR{time step $t$ $\gets$ $1$  to  $\mathbf{t}$ }
        \IF{$Revolute$ $joint$}
        \STATE $\mathbf{T}_{t + 1}$ $\gets$  $\mathbf{Rot}(\hat{\mathbf{q}_1}, \hat{\mathbf{q}_2})$ $\cdot$ $\mathbf{T}_{t}$
        \ELSIF{$Prismatic $ $joint$}
        \STATE $\mathbf{T}_{t + 1}$ $\gets$  $\mathbf{Trans}(\hat{\mathbf{q}_1}, \hat{\mathbf{q}_2})$ $\cdot$ $\mathbf{T}_{t}$
        \ENDIF
        \ENDFOR
        \RETURN $\{\mathbf{T}_t\}_{t=1}^{\mathbf{t}}$
    \end{algorithmic}
\end{algorithm}

We test on 9 objects with diverse appearances. Their categories and part classes are: Box (Lid), Document-Box (Lid), Bucket (Handle), Door (Door), Drawer (Drawer), Notebook (Lid-book), Folder (Lid-book), Lamp (Lid-thin), Clapperboard (Lid-thin).
We show a random view for each object in Fig~\ref{fig:robot_objects}.

The GPS-based heuristic policy is shown in Alg.~\ref{alg:policy}.
To select $\mathbf{G}$, GPS predictions are used for a scoring function $\mathcal{S}$ of grasp proposals. 
For objects with revolute joint, one criterion is the angle between $\{\hat{\mathbf{q}_1}, \hat{\mathbf{q}_2}, \hat{\mathbf{p}}\}$ plane and $\{\hat{\mathbf{q}_1}, \hat{\mathbf{q}_2}, \mathbf{o}\}$ plane, where $\mathbf{o}$ is the position of a grasp.
The angle and the original grasp confidence scores are processed with z-score normalization. The final score is their weighted sum, with the coefficients 1.0, 0.25.
For objects with a prismatic joint, the criterion is the distance from $\mathbf{o}$ to the $\{\hat{\mathbf{q}_1}, \hat{\mathbf{q}_2}, \hat{\mathbf{p}}\}$ plane. The coefficients of the distance and the original grasp confidence scores are 1.0, 0.5.

We also provide robot manipulation videos as an attachment in the supplementary material.

\subsection{Combination with Diffusion Policy}
We conduct a small-scale experiment with our GPS-Policy base on RISE~\cite{wang2024rise}, a diffusion policy model with point cloud input. 
We develop GPS-Policy: we use the trained GPS model to extract GPS prediction for \textit{the initial frame}, and encode them as additional input for RISE.
\textit{For each frame}, the policy predicts future GPS and then uses it as condition to guide action generation.
The task is closing a rotation lid. Observation is recorded via a side-view RGBD camera.
The policy is trained on 5 objects, with 50 demonstrations per object. 
We evaluate the policy on another 5 objects, conducting 10 trials each object under varying poses.
We find that GPS integration boosts the success rate from 32\% to 78\%. 
Notably, GPS-Policy excels at 
contacting the lid at correct position
and closing it via proper path. 
We will extend to more objects, tasks, advanced model and stronger VLA baselines in future work.

\end{document}